\documentclass{article}

\usepackage[preprint]{neurips_2019}

\usepackage[utf8]{inputenc} %
\usepackage[T1]{fontenc}    %
\usepackage{url}            %
\usepackage{booktabs}       %
\usepackage{amsfonts}       %
\usepackage{nicefrac}       %
\usepackage{microtype}      %

\usepackage{bm}             %
\usepackage{amsmath}        %
\usepackage{caption}
\usepackage{comment}
\usepackage{graphicx}
\usepackage{subcaption}
\usepackage{xcolor}
\usepackage[pagebackref=true,breaklinks=true,letterpaper=true,colorlinks,bookmarks=false]{hyperref}
\hypersetup{
    colorlinks,
    linkcolor={red},
    citecolor={green!80!black},
    urlcolor={blue}
}
\usepackage{xspace}
\usepackage[normalem]{ulem}

\def\identical{identical}
\def\id{\identical\xspace}
\def\nonid{non-\identical\xspace}

\def\this{FedAvgM}
\def\thist{\this\xspace}

\def\fedavg{FedAvg}
\def\fedavgt{\fedavg\xspace}

\def\vp{\bm{p}}
\def\vq{\bm{q}}
\def\vw{\bm{w}}
\def\vv{\bm{v}}

\newcommand{\norm}[1]{\left\lVert#1\right\rVert}

\def\dir{\mathsf{Dir}}

\def\eff{\mathrm{eff}}

\newcommand{\algtext}[1]{\texttt{#1}}
\def\thise{$\algtext{\this}$\xspace}
\def\fedavge{$\algtext{\fedavg}$\xspace}

\newcommand{\para}[1]{\left(#1\right)}

\newcommand{\paral}[1]{\left\{#1\right\}}

\newcommand\cut[1]{}

\title{Measuring the Effects of Non-Identical Data Distribution for Federated Visual Classification}
\author{%
  Tzu-Ming Harry Hsu\thanks{work done while interning at Google} \\
  MIT CSAIL \\
  \texttt{stmharry@mit.edu} \\
  \And
  Hang Qi \\
  Google Research\\
  \texttt{hangqi@google.com} \\
  \And
  Matthew Brown \\
  Google Research\\
  \texttt{mtbr@google.com}
}

\begin{document}
\maketitle

\begin{abstract}
Federated Learning enables visual models to be trained in a privacy-preserving way using real-world data from mobile devices. Given their distributed nature, the statistics of the data across these devices is likely to differ significantly. In this work, we look at the effect such non-identical data distributions has on visual classification via Federated Learning. %
We propose a way to synthesize datasets with a continuous range of identicalness and provide performance measures for the Federated Averaging algorithm.
We show that performance degrades as distributions differ more, and propose a mitigation strategy via server momentum. 
Experiments on CIFAR-10 demonstrate improved classification performance over a range of non-identicalness, with classification accuracy improved from 30.1\% to 76.9\% in the most skewed settings.

\end{abstract}

\section{Introduction}
Federated Learning (FL)~\citep{mcmahan2017communication} is a privacy-preserving framework for training models from decentralized user data residing on devices at the edge. With the Federated Averaging algorithm (\fedavge), in each federated learning round, every participating device (also called \emph{client}), receives an initial model from a central server, performs stochastic gradient descent (SGD) on its local dataset and sends back the gradients. The server then aggregates all gradients from the participating clients and updates the starting model. Whilst in data-center training, batches can typically be assumed to be IID (independent and identically distributed), this assumption is unlikely to hold in Federated Learning settings. In this work, we specifically study the effects of non-identical data distributions at each client, assuming the data are drawn independently from differing local distributions. We consider a continuous range of non-identical distributions, and provide empirical results over a range of hyperparameters and optimization strategies.

\section{Related Work}

Several authors have explored the \fedavge algorithm on \nonid client data partitions generated from image classification datasets. \citet{mcmahan2017communication} synthesize pathological \nonid user splits from the MNIST dataset, sorting training examples by class labels and partitioning into shards such that each client is assigned with 2 shards. They demonstrate that \fedavge on \nonid clients still converges to 99\% accuracy, though taking more rounds than \id clients. In a similar sort-and-partition manner, \citet{zhao2018federated} and \citet{ sattler2019robust} generate extreme partitions on the CIFAR-10 dataset, forming a population consisting of 10 clients in total. These settings are somewhat unrealistic, as practical federated learning would typically involve a larger pool of clients, and more complex distributions than simple partitions. 

Other authors look at more realistic data distributions at the client. For example, \citet{caldas2018leaf} use Extended MNIST~\citep{cohen2017emnist} with partitions over writers of the digits, rather than simply partitioning over digit class. Closely related to our work, \cite{yurochkin2019bayesian} use a Dirichlet distribution with concentration parameter 0.5 to synthesize non-identical datasets. We extend this idea, exploring a continuous range of concentrations $\alpha$, with a detailed exploration of optimal hyperparameter and optimization settings.

Prior work on the theoretical side studied the convergence of \fedavge variants under different conditions. \citet{sahu2018convergence} introduce a proximal term to client objectives and prove convergence guarantees. \citet{li2019convergence} analyze \fedavge under proper sampling and averaging schemes in strongly convex problems.

\label{sec:related}
\section{Synthetic Non-Identical Client Data}

In our visual classification task, we assume on every client training examples are drawn independently with class labels following a categorical distribution over $N$ classes parameterized by a vector $\vq$ ($q_i \ge 0, i \in [1, N]$ and $\norm{\vq}_1 = 1$).
To synthesize a population of non-identical clients, we draw $\vq \sim \dir\para{\alpha \vp}$ from a Dirichlet distribution, where $\vp$ characterizes a prior class distribution over $N$ classes, and $\alpha > 0$ is a \emph{concentration} parameter controlling the identicalness among clients. We experiment with 8 values for $\alpha$ to generate populations that cover a spectrum of identicalness. With $\alpha \rightarrow \infty$, all clients have identical distributions to the prior; with $\alpha \rightarrow 0$, on the other extreme, each client holds examples from only one class chosen at random.

\begin{figure}[!h]
    \centering
    \includegraphics[width=\textwidth]{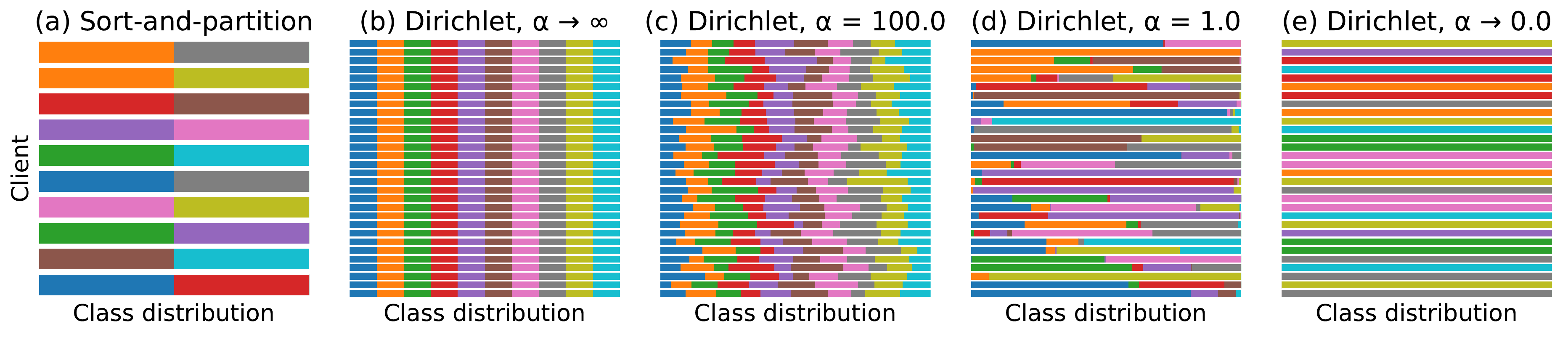}
    \caption{\textbf{Synthetic populations with non-identical clients.} Distribution among classes is represented with different colors. (a) 10 clients generated from the sort-and-partition scheme, each assigned with 2 classes. (b--e) populations generated from Dirichlet distribution with different concentration parameters $\alpha$ respectively, 30 random clients each.}
    \label{fig:dist}
\end{figure}

In this work, we use the CIFAR-10~\citep{krizhevsky2009learning} image classification dataset, which contains 60,000 images (50,000 for training, 10,000 for testing) from 10 classes. We generate balanced populations consisting of 100 clients, each holding 500 images. We set the prior distribution to be uniform across 10 classes, identical to the test set on which we report performance. For every client, given an $\alpha$, we sample $\vq$ and assign the client with the corresponding number of images from 10 classes.
Figure~\ref{fig:dist} illustrates populations drawn from the Dirichlet distribution with different concentration parameters.

\section{Experiments and Results}

Given the above dataset preparation, we now proceed to benchmark the performance of the vanilla \fedavge algorithm across a range of distributions ranging from identical to non-identical.

We use the same CNN architecture and notations as in~\citet{mcmahan2017communication} except that a weight decay of 0.004 is used and \emph{no} learning rate decay schedule is applied. This model is not the state-of-the-art on the CIFAR-10 dataset, but is sufficient to show relative performance for the purposes of our investigation.

\fedavge is run under client batch size $B = 64$, local epoch counts $E \in \paral{1, 5}$, and reporting fraction $C \in \paral{0.05, 0.1, 0.2, 0.4}$ (corresponding to 5, 10, 20, and 40 clients participating in every single round, respectively) for a total of 10,000 communication rounds. We perform hyperparameter search over a grid of client learning rates $\eta \in \paral{10^{-4}, 3 \times 10^{-4}, \ldots, 10^{-1}, 3 \times 10^{-1}}$.

\subsection{Classification Performance with Non-Identical Distributions}

\begin{figure}[!h]
    \centering
    
    \begin{minipage}{.439\textwidth}
        \begin{subfigure}{\textwidth}%
            \phantomsubcaption
            \label{fig:fedavg-alpha:a}
        \end{subfigure}
        \begin{subfigure}{\textwidth}%
            \phantomsubcaption
            \label{fig:fedavg-alpha:b}
        \end{subfigure}
        
        \includegraphics[width=\textwidth]{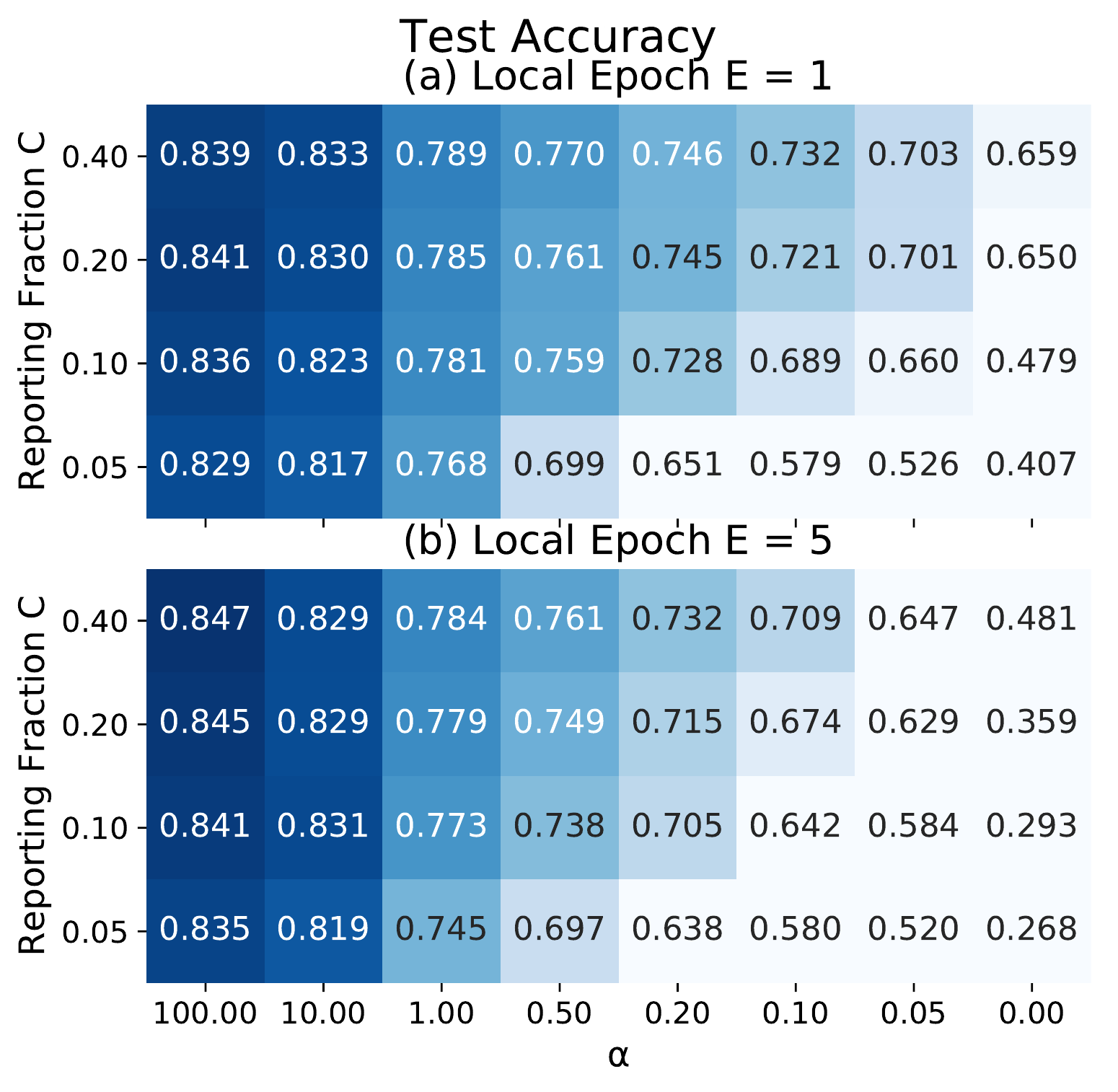}
        \caption{\textbf{\fedavgt accuracy for different $\boldsymbol\alpha$}. 
        Each cell is optimized over learning rates, with each learning rate averaged over 5 runs on different populations under the same $\alpha$.
        }
        \label{fig:fedavg-alpha}
    \end{minipage}%
    \qquad
    \begin{minipage}{.495\textwidth}        
        \includegraphics[width=\textwidth]{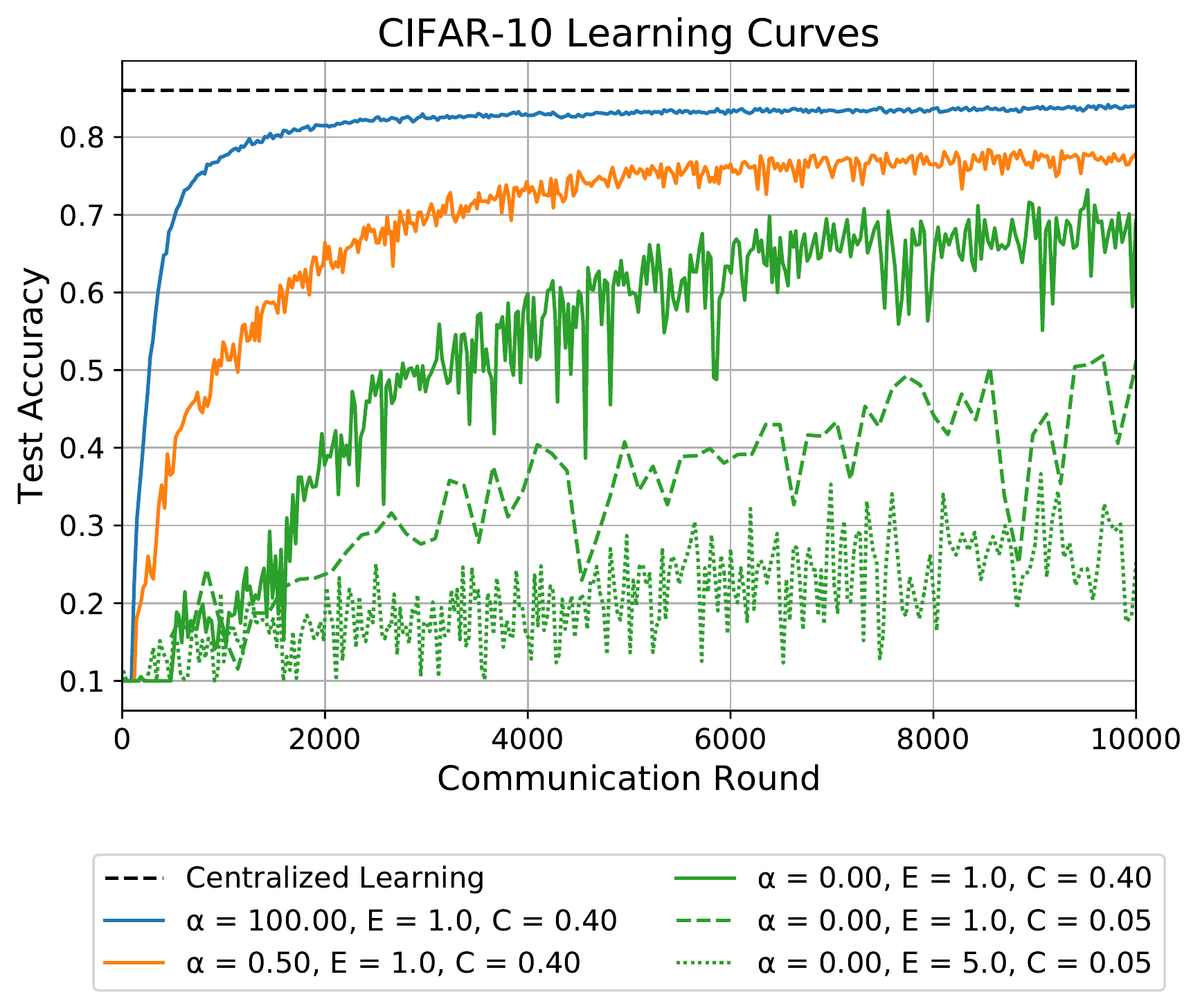}
        \caption{\textbf{\fedavgt learning curves with fixed learning rates.} The centralized learning result (dashed line) is from TensorFlow tutorial~\citep{tensorflowtut}.}
        \label{fig:fedavg-round}
    \end{minipage}%
    
    \vspace{-10pt}
\end{figure}

Figure~\ref{fig:fedavg-alpha} shows classification performance as a function of the Dirichlet concentration parameter $\alpha$ (larger $\alpha$ implies more identical distributions). Significant changes in test accuracy occur around low $\alpha$ when the clients come close to one-class. 
Increasing the reporting fraction $C$ yields diminishing returns, and the gain in performance is especially marginal for identically distributed client datasets. 
Interestingly, for the case of fixed optimization round budget, synchronizing the weights more frequently ($E = 1$) does not always improve the accuracy on non-identical data.

In addition to reduced end-of-training accuracy, we also observe more volatile training error in the case of more non-identical data, see Figure~\ref{fig:fedavg-round}. Runs with small reporting fraction struggle to converge within 10,000 communication rounds.

\begin{figure}[!h]
    \centering
    \begin{subfigure}{\textwidth}%
        \phantomsubcaption
        \label{fig:fedavg-lr:a}
    \end{subfigure}
    \begin{subfigure}{\textwidth}%
        \phantomsubcaption
        \label{fig:fedavg-lr:b}
    \end{subfigure}
    \begin{subfigure}{\textwidth}%
        \phantomsubcaption
        \label{fig:fedavg-lr:c}
    \end{subfigure}
    \begin{subfigure}{\textwidth}%
        \phantomsubcaption
        \label{fig:fedavg-lr:d}
    \end{subfigure}
    
    \includegraphics[width=\textwidth]{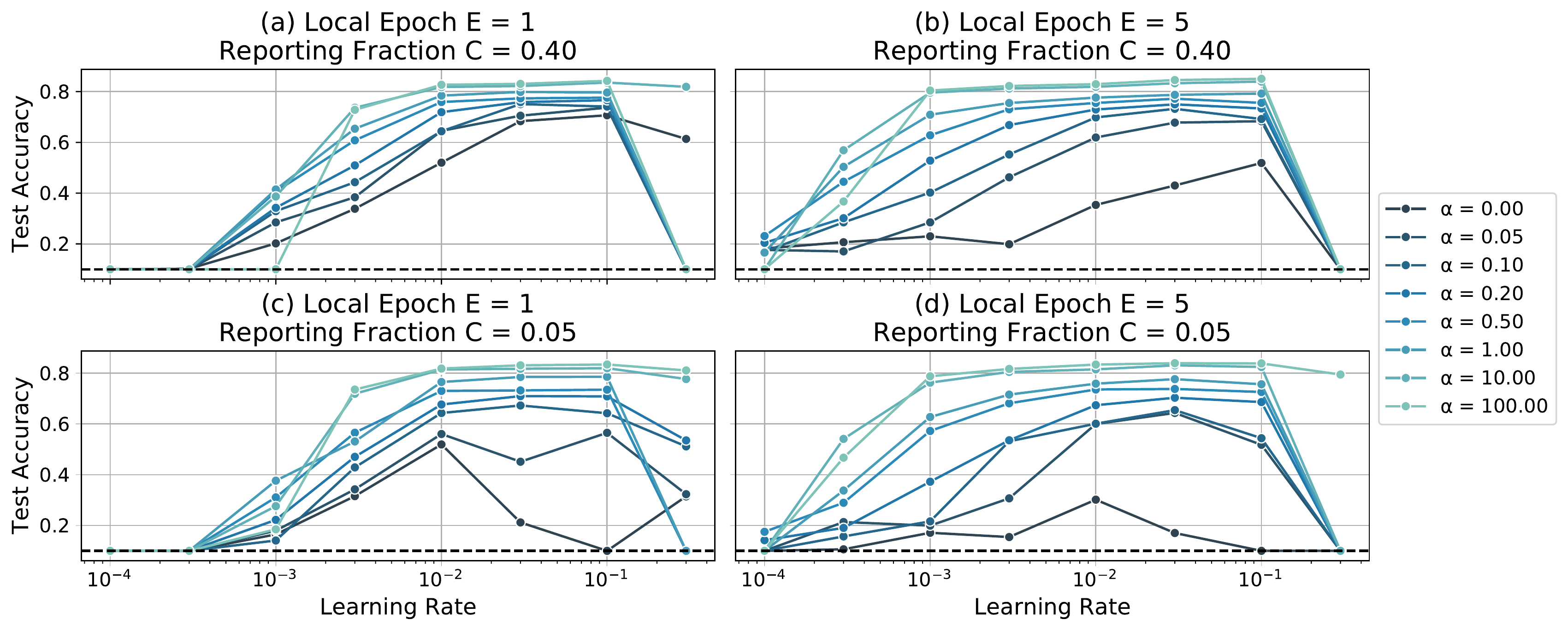}
    \caption{\textbf{\fedavgt test accuracy in hyperparameter search.} (a--b) High and (c--d) low reporting fraction out of 100 clients are demonstrated. Chance accuracy is shown by the dashed line.
    }
    \label{fig:fedavg-lr}
    
    \vspace{-10pt}
\end{figure}

\paragraph{Hyperparameter sensitivity.} 

As well as affecting overall accuracy on the test set, the learning conditions as specified by $C$ and $\alpha$ have a significant effect on hyperparameter sensitivity. On the identical end with large $\alpha$, a range of learning rates (about two orders of magnitude) can produce good accuracy on the test set. However, with smaller values of $C$ and $\alpha$, careful tuning of the learning rate is required to reach good accuracy. See Figure~\ref{fig:fedavg-lr}.

\subsection{Accumulating Model Updates with Momentum}

Using momentum on top of SGD has proven to have great success in accelerating network training by a running accumulation of the gradient history to dampen oscillations. This seems particularly relevant for FL where participating parties may have a sparse distribution of data, and hold a limited subset of labels. In this subsection we test the effect of momentum at the server on the performance of \fedavge.

Vanilla \fedavge updates the weights via $\vw \leftarrow \vw - \Delta \vw$, where $\Delta \vw = \sum_{k=1}^K \frac{n_k}{n} \Delta  \vw_k$ ($n_k$ is the number of examples, $\Delta \vw_k$ is the weight update from $k$'th client, and $n=\sum_{k=1}^{K} n_k$). To add momentum at the server, we instead compute $\vv \leftarrow \beta \vv + \Delta \vw$, and update the model with $\vw \leftarrow \vw - \vv$. We term this approach \thise (Federated Averaging with Server Momentum).

In experiments, we use Nesterov accelerated gradient~\citep{nesterov2007gradient} with momentum $\beta \in \paral{0, 0.7, 0.9, 0.97, 0.99, 0.997}$. The model architecture, client batch size $B$, and learning rate $\eta$ are the same as vanilla \fedavge in the previous subsection. The learning rate of the server optimizer is held constant at 1.0.

\begin{figure}[!h]
    \centering
    \includegraphics[width=\textwidth]{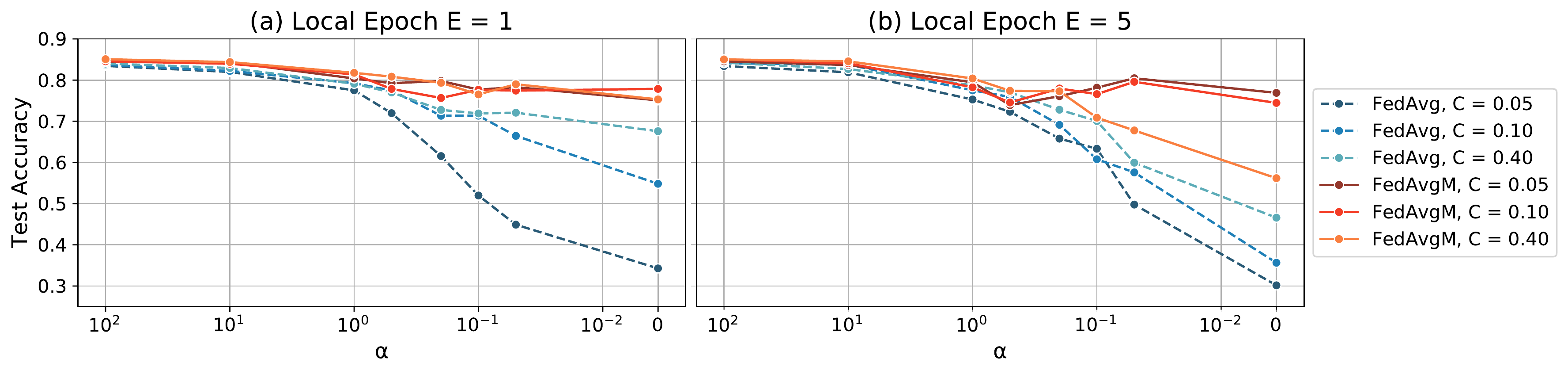}
    \caption{\textbf{\thist and \fedavgt performance curves for different non-identical-ness.} 
    Data is increasingly non-identical to the right. Best viewed in color.
    }
    \label{fig:fedaccum-kl}
    
    \vspace{-10pt}
\end{figure}

\paragraph{Effect of server momentum.}
Figure~\ref{fig:fedaccum-kl} shows the effect of learning with non-identical data both with and without server momentum. The test accuracy improves consistently for \thise over \fedavge, with performance close to the centralized learning baseline ($86.0\%$) in many cases. For example, with $E=1$ and $C=0.05$, \thise performance stays relatively constant and above $75\%$, whereas \fedavge accuracy falls rapidly to around $35\%$.

\begin{figure}[!h]
    \centering
    \includegraphics[width=\textwidth]{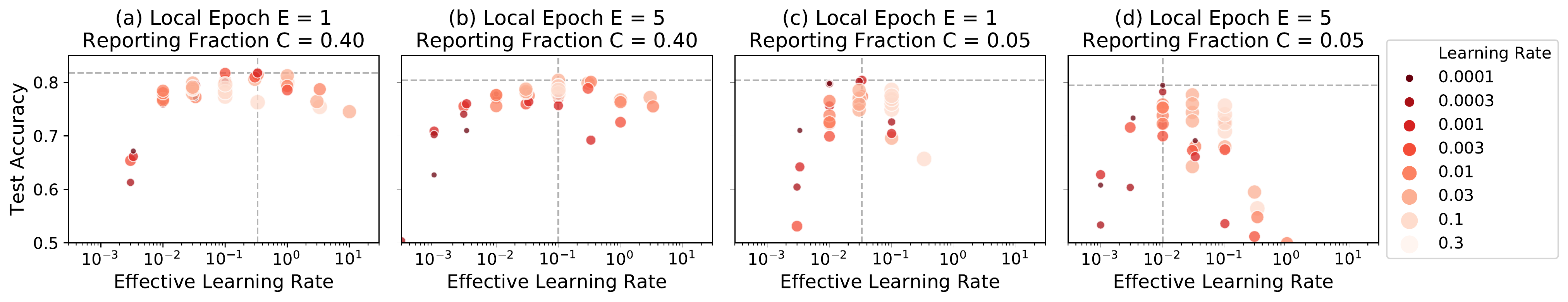}
    \caption{\textbf{Sensitivity of test accuracy for \thist.} Plotted for $\alpha=1$. The effective learning rate is defined as $\eta_\eff = \eta / \para{1 - \beta}$. Sizes are proportional to client learning rate $\eta$ and the most performant point is marked by crosshair.
    }
    \label{fig:fedaccum-lr}
\end{figure}

\paragraph{Hyperparameter dependence on $C$ and $E$.}
Hyperparameter tuning is harder for \thise as it involves an additional hyperparameter $\beta$.
In Figure~\ref{fig:fedaccum-lr}, we plot the accuracy against the effective learning rate defined as $\eta_\eff = \eta / \para{1-\beta}$~\citep{shallue2018measuring} which suggests an optimal $\eta_\eff$ for each set of learning conditions. Notably, when the reporting fraction $C$ is large, the selection of $\eta_\eff$ is easier and a range of values across two orders of magnitude yields reasonable test accuracy. In contrast, when only a few clients are reporting each round, the viable window for $\eta_\eff$ can be as small as just one order of magnitude. To prevent client updates from diverging, we additionally have to use a combination of low absolute learning rate and high momentum. 
The local epoch parameter $E$ affects the choice of learning rate as well. Extensive local optimization increases the variance of clients' weight updates, therefore a lower $\eta_\eff$ is necessary to counteract the noise.

\clearpage
\bibliographystyle{plainnat}
\bibliography{main}

\begin{thebibliography}{12}
\providecommand{\natexlab}[1]{#1}
\providecommand{\url}[1]{\texttt{#1}}
\expandafter\ifx\csname urlstyle\endcsname\relax
  \providecommand{\doi}[1]{doi: #1}\else
  \providecommand{\doi}{doi: \begingroup \urlstyle{rm}\Url}\fi

\bibitem[Caldas et~al.(2018)Caldas, Wu, Li, Kone{\v{c}}n{\`y}, McMahan, Smith,
  and Talwalkar]{caldas2018leaf}
Sebastian Caldas, Peter Wu, Tian Li, Jakub Kone{\v{c}}n{\`y}, H~Brendan
  McMahan, Virginia Smith, and Ameet Talwalkar.
\newblock Leaf: A benchmark for federated settings.
\newblock \emph{arXiv preprint arXiv:1812.01097}, 2018.

\bibitem[Cohen et~al.(2017)Cohen, Afshar, Tapson, and van
  Schaik]{cohen2017emnist}
Gregory Cohen, Saeed Afshar, Jonathan Tapson, and Andr{\'e} van Schaik.
\newblock {EMNIST}: an extension of {MNIST} to handwritten letters.
\newblock \emph{arXiv preprint arXiv:1702.05373}, 2017.

\bibitem[Krizhevsky et~al.(2009)Krizhevsky, Hinton,
  et~al.]{krizhevsky2009learning}
Alex Krizhevsky, Geoffrey Hinton, et~al.
\newblock Learning multiple layers of features from tiny images.
\newblock Technical report, Citeseer, 2009.

\bibitem[Li et~al.(2019)Li, Huang, Yang, Wang, and Zhang]{li2019convergence}
Xiang Li, Kaixuan Huang, Wenhao Yang, Shusen Wang, and Zhihua Zhang.
\newblock On the convergence of {FedAvg} on non-{IID} data.
\newblock \emph{arXiv preprint arXiv:1907.02189}, 2019.

\bibitem[McMahan et~al.(2017)McMahan, Moore, Ramage, Hampson, and
  y~Arcas]{mcmahan2017communication}
Brendan McMahan, Eider Moore, Daniel Ramage, Seth Hampson, and Blaise~Aguera
  y~Arcas.
\newblock Communication-efficient learning of deep networks from decentralized
  data.
\newblock In \emph{Artificial Intelligence and Statistics}, pages 1273--1282,
  2017.

\bibitem[Nesterov(2007)]{nesterov2007gradient}
Yu~Nesterov.
\newblock Gradient methods for minimizing composite objective function.
\newblock 2007.

\bibitem[Sahu et~al.(2018)Sahu, Li, Sanjabi, Zaheer, Talwalkar, and
  Smith]{sahu2018convergence}
Anit~Kumar Sahu, Tian Li, Maziar Sanjabi, Manzil Zaheer, Ameet Talwalkar, and
  Virginia Smith.
\newblock On the convergence of federated optimization in heterogeneous
  networks.
\newblock \emph{arXiv preprint arXiv:1812.06127}, 2018.

\bibitem[Sattler et~al.(2019)Sattler, Wiedemann, M{\"u}ller, and
  Samek]{sattler2019robust}
Felix Sattler, Simon Wiedemann, Klaus-Robert M{\"u}ller, and Wojciech Samek.
\newblock Robust and communication-efficient federated learning from non-{IID}
  data.
\newblock \emph{arXiv preprint arXiv:1903.02891}, 2019.

\bibitem[Shallue et~al.(2018)Shallue, Lee, Antognini, Sohl-Dickstein, Frostig,
  and Dahl]{shallue2018measuring}
Christopher~J Shallue, Jaehoon Lee, Joe Antognini, Jascha Sohl-Dickstein, Roy
  Frostig, and George~E Dahl.
\newblock Measuring the effects of data parallelism on neural network training.
\newblock \emph{arXiv preprint arXiv:1811.03600}, 2018.

\bibitem[TensorFlow()]{tensorflowtut}
TensorFlow.
\newblock Advanced convolutional neural networks.
\newblock URL \url{https://www.tensorflow.org/tutorials/images/deep_cnn}.

\bibitem[Yurochkin et~al.(2019)Yurochkin, Agarwal, Ghosh, Greenewald, Hoang,
  and Khazaeni]{yurochkin2019bayesian}
Mikhail Yurochkin, Mayank Agarwal, Soumya Ghosh, Kristjan Greenewald, Nghia
  Hoang, and Yasaman Khazaeni.
\newblock Bayesian nonparametric federated learning of neural networks.
\newblock In \emph{International Conference on Machine Learning}, pages
  7252--7261, 2019.

\bibitem[Zhao et~al.(2018)Zhao, Li, Lai, Suda, Civin, and
  Chandra]{zhao2018federated}
Yue Zhao, Meng Li, Liangzhen Lai, Naveen Suda, Damon Civin, and Vikas Chandra.
\newblock Federated learning with non-{IID} data.
\newblock \emph{arXiv preprint arXiv:1806.00582}, 2018.

\end{thebibliography}
\end{document}